\documentclass[sigconf]
{acmart}

\usepackage{colortbl}
\usepackage{bbding}
\usepackage{tabularx}
\usepackage[normalem]{ulem}
\usepackage{amsmath,amsfonts}
\usepackage{algorithmic}
\usepackage{graphicx}
\usepackage{textcomp}
\usepackage{multicol}
\usepackage{caption}
\usepackage{xspace}
\usepackage{multirow}
\usepackage{booktabs}
\usepackage{bm}
\usepackage{xcolor}
\usepackage{makecell}

\usepackage{CJKutf8}

\newcommand{\eg}{\emph{e.g.,}\xspace}
\newcommand{\ie}{\emph{i.e.,}\xspace}
\definecolor{light-gray}{gray}{0.9}
\AtBeginDocument{%
  \providecommand\BibTeX{{%
    \normalfont B\kern-0.5em{\scshape i\kern-0.25em b}\kern-0.8em\TeX}}}

\copyrightyear{2024}
\acmYear{2024}
\setcopyright{acmlicensed}\acmConference[KDD '24]{Proceedings of the 30th ACM SIGKDD Conference on Knowledge Discovery and Data Mining}{August 25--29, 2024}{Barcelona, Spain}
\acmBooktitle{Proceedings of the 30th ACM SIGKDD Conference on Knowledge Discovery and Data Mining (KDD '24), August 25--29, 2024, Barcelona, Spain} \acmDOI{10.1145/3637528.3671583} \acmISBN{979-8-4007-0490-1/24/08}
\begin{document}

\title{Know Your Needs Better: Towards Structured Understanding of\\ Marketer Demands with Analogical Reasoning Augmented LLMs}

\author{Junjie Wang}
\authornote{Equal contributions}
\affiliation{%
  \institution{Zhejiang University}
  \institution{Ant Group}
  \city{Hangzhou}
  \country{China}
}
\email{wangjj2018@zju.edu.cn}

\author{Dan Yang}
\authornotemark[1]
\affiliation{%
  \institution{Ant Group}
  \city{Hangzhou}
  \country{China}
}
\email{luoyin.yd@antgroup.com}

\author{Binbin Hu}
\authornotemark[1]
\affiliation{%
  \institution{Ant Group}
  \city{Hangzhou}
  \country{China}
}
\email{bin.hbb@antfin.com}

\author{Yue Shen}
\affiliation{%
  \institution{Ant Group}
  \city{Hangzhou}
  \country{China}
}
\email{zhanying@antfin.com}


\author{Wen Zhang}
\authornote{Corresponding author}
\affiliation{%
  \institution{Zhejiang University}
  \city{Hangzhou}
  \country{China}
}
\email{zhang.wen@zju.edu.cn}

\author{Jinjie Gu}
\authornotemark[2]
\affiliation{%
  \institution{Ant Group}
  \city{Hangzhou}
  \country{China}
}
\email{jinjie.gujj@antfin.com}

\renewcommand{\shortauthors}{Wang, Yang and Hu, et al.}

\begin{abstract}

In this paper, we explore a new way for user targeting, where non-expert marketers could select their target users solely given demands in natural language form. The key to this issue is how to transform natural languages into practical structured logical languages, i.e., the structured understanding of marketer demands. In practical scenarios, the demands of non-expert marketers are often abstract and diverse. Considering the impressive natural language processing ability of large language models (LLMs), we try to leverage LLMs to solve this issue. To stimulate the LLMs' reasoning ability, the chain-of-thought (CoT) prompting method is widely used, but existing methods still have some limitations in our scenario: (1) Previous methods either use simple ``Let's think step by step'' spells or provide fixed examples in demonstrations without considering compatibility between prompts and concrete questions, making LLMs ineffective when the marketers' demands are abstract and diverse. (2) Previous methods are often implemented in closed-source models or excessively large models, which is not suitable in industrial practical scenarios. Based on these, we propose ARALLM (\ie Analogical Reasoning Augmented Large Language Models) consisting of two modules: Analogical Reasoning based Prompting and Reasoning-Augmented Multi-Task Model Distillation. 
To be specific, we first construct a reasoning library consisting of several high-quality and topic-rich reasoning examples. 
Then, we adopt a retrieval-based method to conduct analogical reasoning with the help of the reasoning library. The experimental results show that this prompting strategy achieves better performance than the ordinary prompting method. Beyond that, we distill knowledge from super LLMs (GPT-3.5) to fine-tune smaller student LLMs in a multi-task training paradigm, enabling the models to be easily deployed in practical environments. Part of our data and code can be found at \url{https://github.com/alipay/Analogic-Reasoning-Augmented-Large-Language-Model }.
\end{abstract}





\maketitle

\begin{figure*}[h]
    \centering
    \includegraphics[width=\textwidth]{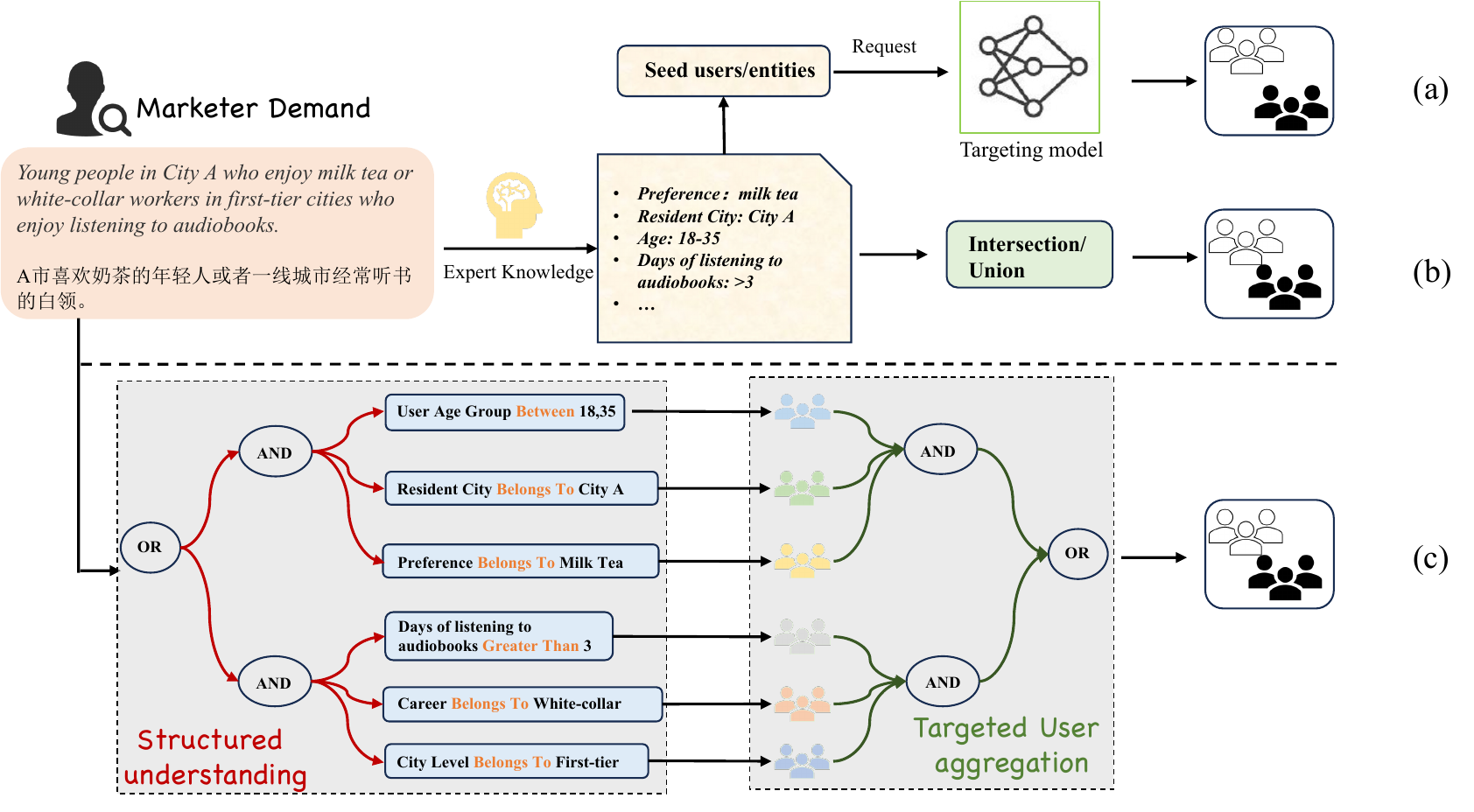}
    \caption{The comparison of user targeting approach.}
    \label{comparison}
\end{figure*}

\section{Introduction}

Recently, the practice of user targeting has gained significant attention in real-world applications (\eg Alipay and WeChat), as highlighted by a variety of studies~\cite{yang2023would,mangalampalli2011feature,shen2015effective,zhuang2020hubble,zhu2021learning}. This approach has the excellent potential to attract high-quality users for specific campaigns, aligning with the goals of marketers to enhance conversions and minimize operational expenses.
Roughly speaking, current methods devoted to user targeting
mainly fall into two lines: \emph{model-based methods} (Figure \ref{comparison} (a)) that perform expansion of seed users/entities with well-designed neural networks~\cite{yang2023would,zhuang2020hubble,zhu2021learning} and \emph{rule-based methods} (Figure \ref{comparison} (b)) that manually group users with different tags based on domain knowledge~\cite{mangalampalli2011feature,shen2015effective}.

Unfortunately, both of the above prior studies have primarily emphasized the intricate architectures for effectively and efficiently gathering targeted users, while commonly ignoring the natural and significant gap between marketers' demand and the capability of current models. 
In particular, current user targeting pipelines mainly force the marketers to decompose their demands into multiple tags/entities, \eg ``Preference'', ``Resident City'' and ``User Age Group'' which are characterized in Figure \ref{comparison}. The process incurs several weaknesses: 
i) unfriendly to marketers;
ii) time-consuming for manual deconstruction of demands;
and iii) unpromising performance due to partial tags/entities.

Therefore, in this paper, we take the initial stride towards enhancing current user targeting systems by probing into the structured understanding of marketers' demands in an automatic way, \emph{enabling the system to know your needs better solely based on natural language inputs}.
As shown in Figure \ref{comparison} (c), 
the marketers only need to input their native demands like ``Young people in City A who enjoy milk tea or white-collar workers in first-tier cities who enjoy listening to audiobooks'' and our new approach can instantly conduct structured decomposition of marketers' demands and thus provide the user targeting cards. After verifying the cards, they can export the target users with just one click and the system will automatically run the existing targeted user aggregation modules. 

\textbf{The core of this new approach lies in how to transform natural languages into structured forms}, or so-called \textit{structured understanding of marketer demands}. To address this issue, we first draw upon existing logical and programming languages to design a logical expression that offers both readability and practical applicability for the structured representation of marketers' demands. We name it as SELL, \ie \underline{S}tructured and \underline{E}ditable \underline{L}ogical \underline{L}anguage, which mainly consists of Keys (\eg Resident City), Values (\eg City A), Operators (\eg Belongs To) and Intersection/Union symbols (\ie AND, OR). It forms the basis of the user targeting cards shown in Figure \ref{comparison}(c). The main concern of our work is transforming the Natural Language into SELL (\textbf{NL2SELL}). 
For example, the natural language ``Young people in City A who enjoy milk tea or white-collar workers in first-tier cities who enjoy listening to audiobook'' can be translated into SELL as ``((\textit{Resident City\#Belongs To\# City A}) \textit{AND} (\textit{User Age Group\#Between\#18,35}) \textit{AND} (\textit{Preference\#Belongs To\#Milk Tea})) \textit{OR} ((\textit{City Level\#Belongs To\#First-tier}) \textit{AND} (\textit{Days of listening to audiobooks\#Greater Than\#3}) \textit{AND} (\textit{Career\#Belongs To\#White-collar}))''. 

Previous research ~\cite{Liu_Hu_Wen_Yu_2023,gu2023few,yang2023harnessing,lu-etal-2022-parsing} has highlighted the remarkable abilities of Large Language Models (LLMs) in language translation tasks. LLMs are renowned for their robust natural language processing and zero-shot capabilities, which render them possibly effective in NL2SELL. Consequently, we opt to employ LLMs to tackle this task.
However, there are still some challenges in using LLMs in our scenario:
\textbf{(1) The challenge of reasoning accuracy.} In practical user targeting scenarios, understanding of marketer's demands could be very challenging. For example, a marketer might want to market products related to education, targeting parents of middle school students who are focused on education. Therefore, the demand she inputs into the system is a simple sentence ``Parents of middle school students'', and we need to use existing tags to convert this demand into a structured expression, such as ``\textit{(Marital Status\#Belongs\#True) AND (User Child Age\#Between\#12,15) AND (Preference\#Belongs To\#Education)}''. However, the key ``Marital Status, User Child Age, and Preferences'' do not directly appear in the marketer's demand, which requires the LLMs to possess a high level of language comprehension and reasoning abilities. Recently, it \cite{wei2022chain,kojima2022large} has been pointed out that prompting LLMs with chain-of-thought (CoT) could enhance the reasoning ability of LLMs. These methods either provide fixed reasoning examples as demonstrations through few-shot learning or tell the model ``Let's think step by step'' for zero-shot learning. Though effective, they ignore the compatibility between prompts and specific questions. Considering the diversity of marketers' demands themselves, zero-shot learning or having fixed reasoning examples in few-shot learning may not effectively stimulate the LLMs' reasoning ability \cite{rubin2021learning}. Ideally, we can customize a reasoning process for each question to induce the model to reason better, but it is time-consuming and laborious. 
\textbf{(2) The challenge of reasoning speed and resource consumption.} Although the reasoning ability of super-large language models can be developed through prompt engineering, it is unacceptable to deploy excessively huge LLMs or API-based LLMs in practical situations. For extremely huge LLMs, excessive parameters will consume a lot of resources, and the reasoning speed is also slower. For API-based LLMs such as GPT-3.5, both the security of private data and the cost of API calls are thorny issues. Our model needs to be deployed in real ToC (To Customer) scenarios, where it will generate at least hundreds of tasks daily. Therefore, there is an urgent requirement to increase the inference speed, reduce the cost of inference, and ensure data security. Thus, smaller, free, and white-box LLMs are preferred.

To address these challenges, we propose a framework called \underline{A}nalogic \underline{R}easoning \underline{A}ugmented  \underline{L}arge \underline{L}anguage \underline{M}odel (ARALLM) to tackle the NL2SELL task, which consists of two modules: Analogical Reasoning based Prompting and Reasoning-Augmented Multi-Task Model Distillation.
\textbf{(1) Analogical Reasoning based Prompting.} Creating a CoT for each new targeting demand is impractical and time-consuming (considering the need for deployment). Thus, we propose Analogical Reasoning based Prompting.
The fundamental concept of it is that when two targeting demands are similar, they might follow similar reasoning steps and have similar conditional expression structures. Therefore, we construct a compact yet reliable reasoning library to serve as a reference for numerous demands that have unknown reasoning steps. This approach helps reduce the effort of manually writing reasoning steps for every demand. Additionally, we can offer assistance for any unfamiliar demands by employing analogical reasoning with retrieval.
\textbf{(2) Reasoning-Augmented Multi-Task Model Distillation.} Since the white-box and smaller model is preferred, we distill knowledge from teacher LLMs (GPT-3.5) to construct an NL2SELL dataset with over 10,000 pieces of training samples and then propose a novel multi-task training method with reasoning augmentation to improve the reasoning ability of student LLMs.

We summarize our contributions as follows:
(1) We explore a new way for marketer-friendly user targeting, which focuses on the structured understanding of marketers' raw demands.
(2) We propose an innovative framework called ARALLM, which is composed of the Analogical Reasoning based Prompting module and the Reasoning-Augmented Multi-Task Distillation module, to address the above task.
(3) We demonstrate the superiority of the proposed ARALLM framework through extensive experiments on real-world datasets and make this LLMs-based framework operate in real industry scenarios.

\section{Method}
\begin{figure*}[t]
    \centering
    \includegraphics[width=0.8\textwidth]{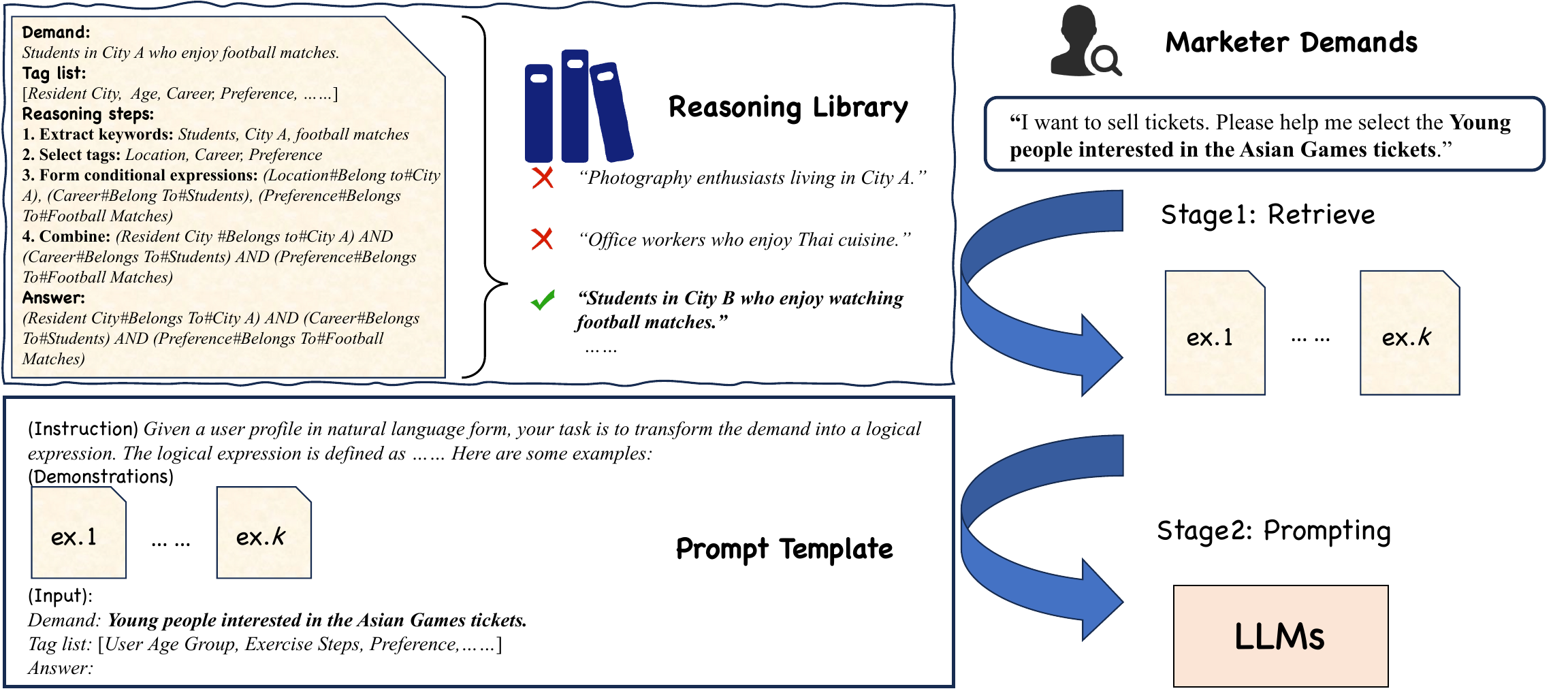}
    \caption{Overview of analogical reasoning based prompting methods.}
    \label{analogical overview}
\end{figure*}
In this work, we \textbf{focus on how to convert a sentence of marketer's demand $d$ into a standard structured expression SELL}, while how to parse and execute the {SELL} and return the target users from the database will be handled by the internal system, which does not constitute the core of this article. Inspired by the analogical reasoning \cite{yao2023analogical,turney2008latent,gentner1983structure} in the classic artificial intelligence field and CoT prompting methods in LLMs, we first propose a novel prompting method enhanced by analogical reasoning and then adopt a reasoning-augmented multi-task training strategy to fine-tune smaller LLMs.
\subsection{Design of SELL}
We first explain the design of SELL to help readers understand our task better.
The SELL is mainly composed of four elements: \textbf{Keys}, \textbf{Values}, \textbf{Operators}, and \textbf{Intersection and Union symbols}:

(1) \textbf{Keys} are a series of tags that describe the features of the user, such as \textit{Gender}, \textit{Monthly Income}, \textit{Pet Owning}, and so on.

(2) \textbf{Values} are the fillings of the corresponding keys. They can be generally divided into three types: numerical type, string type, and boolean type. 

(3) \textbf{Operators} represent the relationship between keys and values. For numerical values, seven types of operators can be used to connect them with corresponding keys: \textit{Equal To, Greater Than, Less Than, Not Equal To, Not Greater Than, Not Less Than,} and \textit{Between}. For values of string type and boolean type, there are two types of operators: \textit{Belongs To} and \textit{Not Belongs To}. Keys, operators, and values can form a basic \textbf{conditional expression} in {SELL}, formatted as ``(\textit{key\#operator\#value})'', such as ``(\textit{User Marital Status\#Belongs\#True})''. A conditional expression represents a cluster of target users.

(4) \textbf{Intersection and Union symbols} are used to take the intersection or union of multiple conditional expressions (targeted users). We use the symbols \textit{AND} and \textit{OR} to denote the intersection and union operations, respectively.

Compared to the popular database languages SQL and logic language FOL, the advantages of SELL are: \textbf{(1)} SELL is simpler in expression and has much fewer syntax symbols, making it easier for non-programming experts to understand. \textbf{(2)} Since SELL can be transformed into a clear tree structure, marketers can easily modify them even if the generated results have flaws. Due to space limitations, we recommend readers to read more details about SELL in the supplementary materials.

\subsection{Prompting With Analogical Reasoning}
The core idea of analogical reasoning in the NL2SELL task is that if two demands are similar, they may share similar reasoning steps, as well as similar logical structures. Thus, we construct a small but high-quality reasoning library to provide references for numerous demands with unknown reasoning steps, minimizing the cost of manually writing reasoning steps for each of them. 

\subsubsection{Construction of Reasoning Library}
To construct a reasoning library, we first randomly collect $\mathcal{N}_{RL}$ marketer demands from real-world scenarios
and write their corresponding answers based on our expert knowledge to form Q\&A pairs $\mathcal{R} = \{(d_{e},s_{e})\}_{e=1}^{\mathcal{N}_{RL}}$, where $d_{e}$ is the demand, $s_{e}$ is the corresponding SELL expressions. The collected marketer's demands encompass services that are popular on Alipay, including education, technology, reading, travel, financial management, insurance, government affairs, and public welfare, thus ensuring that the reasoning library constructed based on this can cover as many demand topics as possible. Then we use a small amount of manual effort to write corresponding reasoning details for 10\% random selected Q\&A pairs $\in$ $\mathcal{R}$ to construct the seed examples in the format $(d_{e},r_{e}, s_{e})$, where $r_{e}$ represent the reasoning details. We generally summarize the procedure of solving the NL2SELL problem as four steps:
\begin{enumerate}
    \item Extract keywords: Extract the keywords from the demand.
    \item Select tags: Select the most relevant tags from the tag list based on the keyword as the key in SELL.
    \item Form conditional expressions: Based on the selected keywords and contextual information, fill in the corresponding operators and values to form conditional expressions.
    \item Combine: Combine conditional expressions using intersection or union symbols.
\end{enumerate}
What we need to do is to fill in the details of these four steps.

For the remaining data in $\mathcal{R}$, we utilize the OpenAI gpt-3.5-turbo-1106 (GPT-3.5) API to help us fill the corresponding reasoning steps, and our seed examples serve as demonstrations in the prompt. Since we have standardized the overall framework of reasoning, it is not difficult for GPT-3.5 to complete this part of the reasoning steps, cases are shown in supplemental materials. After being verified by experts. we harvest a batch of solid reasoning details to form a reasoning library $\mathcal{R} = \{(d_{e},r_{e},s_{e})\}_{e=1}^{\mathcal{N}_{RL}}$.
\subsubsection{Retrieval-Based Analogical Prompt Construction}
After the construction of the reasoning library, we can provide references for unknown demands through analogical reasoning. As shown in Figure \ref{analogical overview}, given a marketer demand $d_{p}$ that needs to predict the corresponding SELL expression, we first retrieve similar demands $d_{e}$ in $\mathcal{R}$ based on their text embedding similarity: 
\begin{equation}
    sim(d_{p},d_{e}) = cos(\textbf{E}(d_{p}),\textbf{E}(d_{e})), d_{e}\in \mathcal{R}
\end{equation}
where \textbf{E} is the embedding model in which we use BGE-large-zh \cite{bge_embedding,llm_embedder} off the shelf as it is well-performed in the Chinese text retrieval field. The top $k$ demands in $\mathcal{R}$ that get the highest similarity score when compared with $d_{p}$ will be fetched as analogical examples in this stage, along with their corresponding $r$ and $s$.

In addition, to help the LLMs better organize answers, we should also inform the LLMs of the tags in the marketing database as explicit knowledge. To address the issue of an excessive number of tags, we still use BGE as a retriever to retrieve the most relevant \textit{n} tags of every demand from the marketing database to construct a part of the prompt.
The top \textit{n} tags that get the highest similarity scores when compared with demand will form a small-scale tag list $T = \{t_x\}_{x=1}^n$ to provide a reference for selecting keys in SELL.

Therefore, the basic organizational form of the final prompt $x_p$ for predicting is :
\begin{equation}
   x_p = concat(\mathcal{I},\{(d_{e},T_{e},r_{e},s_{e})\}_{e=1}^k,(d_{p},T_{p}))
   \label{prompt}
\end{equation}

where $\mathcal{I}$ is the instruction which simply describes the task, \textit{k} is the number of analogical examples, $T_{e}$ and $T_{p}$ are the most relevant tag lists of the $d_{e}$ and $d_{p}$ respectively. 

Compared to the ordinary CoT method, our approach exhibits several main differences:
(1) The NL2SELL task is more challenging. Unlike solving mathematical problems or common sense reasoning tasks, the training data for LLMs don't inherently include the syntax of SELL. This highlights the importance of appropriate reasoning steps.
(2) We place a greater emphasis on the compatibility between prompts and questions. We believe that using an analogical reasoning based prompting method can make LLMs more adaptable to specific problems, rather than relying on fixed few-shot examples, which could potentially limit the model's performance.

\subsection{Reasoning-Augmented Multi-Task Model Distillation}
To train a private small model for deployment, we first distill knowledge from teacher LLM (GPT-3.5) to construct an NL2SELL dataset with over 10,000 pieces of training samples and then propose a multi-task training method with reasoning augmentation.
\subsubsection{Knowledge Distillation}
\label{distill}
We collect nearly 10,000 demands from our daily marketing environment. Following the analogical reasoning method mentioned above, we construct input prompts for each demand in the form of Equation (\ref{prompt}), where analogical examples come from the previously constructed reasoning library. Subsequently, we call the GPT-3.5 API to generate SELL answers based on the above prompts, which will then be further verified and corrected through manual crowdsourcing.
We denote this part of data comes from \textit{demand to answer} paradigm.

It can be seen that the above distillation method relies on \textit{a large number of native demands and tedious manual calibration}. At the same time, due to the different frequencies of user tags used in actual marketing activities, the distilled data has \textit{a long tail distribution}. This means that some demands based on unpopular user tags only exist in a small amount of data, resulting in poor response to such demands. To address the above two issues, we propose a distillation approach based on \textit{answer to demand}, which can be further divided into two stages: \textbf{answer generation} and \textbf{demand generation}.

$\bullet$ Answer generation: We first construct a series of logical templates of SELL. The content of logical templates is obtained from the marketing database through random sampling. Specifically, we first randomly sample a tag from the database as a key. Then we sample its operator based on the type of keys. Finally, we randomly select a value from all possible values of that key, thus constructing a unit conditional expression in the form of (\textit{key}\#\textit{operator}\#\textit{value}). Different sampled conditional expressions can be combined by union and intersection to form different types of targeted users. To ensure that the synthesized answers are logically reasonable, we will limit the number of conditional expressions in the logical template, and abandon overly complex intersection and union operations.

$\bullet$ Demand generation:
Considering the strong ability of GPT-3.5, we provide some ($d_{e},s_{e}$) examples in 
$\mathcal{R}$ as demonstrations to guide GPT-3.5 to write demands given above synthesized answers:
\begin{equation}
concat((d_{e},s_{e}),s_{i})\xrightarrow{GPT-3.5}d_{i}
\end{equation}
where $s_{i}$ is the synthesized SELL answers. The language fluency and logical rationality of the generated demand will be checked.

We denote this part of data distilled through the \textit{answer to demand} pipeline. We obtain 1,200 pieces of data in this way and combine them with data coming from the \textit{demand to answer} pipeline to form the final training data $\mathcal{D}_{train} = \{(d_{i}, s_{i})\}_{i=1}^N$ for small model, where \textit{N} is the number of training data.
\subsubsection{Reasoning-Augmented Multi-Task Fine-tune}
\label{fine-tune}
The training objective of the normal distillation is to minimize the loss \textit{L}:
\begin{equation}
    L = \frac{1}{N}\sum_{i=1}^{N}CE(f(d_{i}),s_{i}), (d_i,s_i) \in \mathcal{D}_{train}
\end{equation}
where \textit{f} represents the model, $CE$ is the cross-entropy loss function.

It \cite{hsieh2023distilling,kim2023cot} has been pointed out that fine-tuning LLMs using data combined with reasoning steps can enhance the reasoning capabilities of the LLMs. Inspired by \cite{hsieh2023distilling}, we first prompt GPT-3.5 to generate intermediate reasoning steps of $\mathcal{D}_{train}$ based on $\mathcal{R}$:
\begin{equation}
    concat(({d}_{e},r_{e},s_{e}),(d_{i},s_{i}))\xrightarrow{GPT-3.5}r_{i}
\end{equation}
where $(d_{e},r_{e},s_{e}) \in \mathcal{R}$,
$(d_{i}, s_{i}) \in \mathcal{D}_{train}$, $r_{i}$ is the corresponding generated reasoning steps of $(d_{i},s_{i})$. Subsequently, we define a multi-task training strategy in which not only $s_{i}$ but also $r_{i}$ are used as supervised labels to train the model.

To be specific, we denote the input of these two tasks as:
\begin{equation}
    x_{i_s} = concat(\mathcal{I}_s,\{(d_{e},T_{e},r_{e},s_{e})\}_{e=1}^k,(d_{i},T_{i}))
    \label{s-input}
\end{equation}
\begin{equation}
    x_{i_r} = concat(\mathcal{I}_r,\{(d_{e},T_{e},r_{e},s_{e})\}_{e=1}^k,(d_{i},T_{i}))
    \label{r-input}
\end{equation}
where the only difference between them is the content of instruction $\mathcal{I}_s$ and $\mathcal{I}_r$ which guide the model to generate the SELL answer or reasoning steps respectively. The corresponding training objectives of these two tasks are:
\begin{equation}
    L_s = \frac{1}{N}\sum_{i=1}^{N}CE(f(x_{i_s}),s_{i})
    \label{l-s}
\end{equation}
\begin{equation}
    L_r = \frac{1}{N}\sum_{i=1}^{N}CE(f(x_{i_r}),r_{i})
    \label{l-r}
\end{equation}

The final loss function of multi-task fine-tuning is defined as:
\begin{equation}
    L_{total} = L_s + L_r
\end{equation}

\section{Experiments}
\begin{table}[t]
  \caption{Data statistics. Number(\textit{d}), MaxLen(\textit{d}), MinLen(\textit{d}) and AvgLen(\textit{d}) denote the number, the maximum string length, the minimum string length and the average string length of the demands.}
  \label{datasets}
  \begin{tabular}{c|c|c|c|c}
    \toprule
     & Number(\textit{d}) & MaxLen(\textit{d})  & MinLen(\textit{d}) & AvgLen(\textit{d})\\
    \midrule
    $\mathcal{R}$ & 150 & 42 & 7 & 16.5\\
    $\mathcal{D}_{train}$ & 11329 & 207 & 1 & 16.8\\
    $\mathcal{D}_{test}$& 170 & 48 & 3 & 17.0\\
  \bottomrule
  \end{tabular}
\end{table}

\subsection{Experimental Settings}
\subsubsection{Benchmarks} We manually construct an NL2SELL testing benchmark for predicting, which contains 170 pieces of data based on expert knowledge in the format $\mathcal{D}_{test} = \{(d_{p},s_{p})\}_{p=1}^{170}$. The demands in $\mathcal{D}_{test}$ do not exist in the training set $\mathcal{D}_{train}$ or reasoning library $\mathcal{R}$, eliminating the issue of label leakage. Table \ref{datasets} shows the statistics of these 3 parts of data\footnote{The dataset does not contain any Personal Identifiable Information (PII). The dataset is desensitized and encrypted. Adequate data protection was carried out during the experiment to prevent the risk of data copy leakage.}. It should be noted that all training data generated by GPT-3.5 are strictly verified by well-trained volunteers, and we provide some examples in the supplementary materials. We will open source all these datasets on request.
\subsubsection{Baseline}We first use GPT-3.5 to test the effectiveness of the analogical reasoning based prompting method. For comparison, we propose four additional variants of prompts as baselines:

    $\bullet$ Zero-shot: Only given the instruction $\mathcal{I}$, tag list $T_p$ and a test demand $d_{p}$.
    
    $\bullet$ Fixed few-shot: Given the instruction $\mathcal{I}$, \textit{k} fixed examples $\{(d_e,s_e)\}_{e=1}^k \in \mathcal{R}$ for all $d_{p} \in \mathcal{D}_{test}$, tag list $T_p$ and a $d_{p}$.
    
    $\bullet$ Fixed few-shot + RS (Reasoning Steps): Given the instruction $\mathcal{I}$, \textit{k} fixed examples $\{(d_e,r_e,s_e)\}_{e=1}^k \in \mathcal{R}$ for all $d_{p} \in \mathcal{D}_{test}$, tag list $T_p$ and a $d_{p}$.
    
    $\bullet$ Random few-shot + RS: Given the instruction $\mathcal{I}$, \textit{k} randomly sampled examples $\{(d_e,r_e,s_e)\}_{e=1}^k \in \mathcal{R}$ for every $d_{p} \in \mathcal{D}_{test}$ ,tag list $T_p$ and a $d_{p}$. Different $d_{p}$ has different randomly sampled examples.

In the knowledge distillation stage, we choose ChatGLM2-6B-32K \cite{du2022glm} and Baichuan2-13B-Chat \cite{baichuan2023baichuan2} as student LLMs, considering their better comprehension ability in dealing with long Chinese texts compared to other models. In addition to fine-tuning methods mentioned in Section \ref{distill}, we also propose three other training strategies as baselines:

    $\bullet$ - MT: Training LLMs without multi-task (MT) strategy. The model input is (\ref{s-input}) and the training objective is (\ref{l-s}).
    
    $\bullet$ - RS: Training LLMs with multi-task strategy but the reasoning steps $r_e$ of examples in (\ref{s-input}) and (\ref{r-input}) are not given in the inputs, in which way we can explore the impact of analogical reasoning.
    
    $\bullet$ Normal: Training LLMs in a most normal way, in which the input of the model is the simple combination of instruction $\mathcal{I}$ and a demand $d_{i}$, and the supervised label is $s_{i}$ only.

\subsubsection{Evaluation metrics}
To conduct a more reasonable and comprehensive evaluation, we evaluate the output of the LLMs on the NL2SELL task from 4 perspectives:

    $\bullet$ Structure accuracy (Structure Acc.): The logic structure's accuracy of the predicted SELL expression. We remove the keys, operators, and values in the SELL, leaving only intersection and union symbols and separators as logic structures. For example, the structure of ``((\textit{Location}\#\textit{Belongs To}\#\textit{Hangzhou}) \textit{OR} (\textit{Location\#Belongs To\#Shanghai})) \textit{AND} (\textit{Pet Owning\#Belongs To\#True})'' is ``((\#\#) \textit{OR} (\#\#)) \textit{AND} (\#\#)''. We use the mean value of the Levenshtein distance (L) \cite{levenshtein1966binary} and the Ratcliff/Obershelp similarity (R/O) \cite{black2004ratcliff} to evaluate the structure accuracy compared to label SELL.
    
    $\bullet$ Overall accuracy (Overall Acc.): We use the ScareBLEU \cite{sacrebleu} score to evaluate the overall similarity of the output and label texts as the overall accuracy. We set this as the main metric to evaluate the quality of predicted SELL answers.
    
    $\bullet$ Human evaluation (Human Eval.): We conduct more fine-grained evaluations through human evaluation. Annotators who receive good training will score the accuracy of the key, value, and logic between 0 and 10, and finally provide an overall score. 
    
    $\bullet$ GPT4 evaluation (GPT4 Eval.): We use the most advanced LLM gpt-4-turbo-preview (GPT-4) as the judges. To be specific, we prompt the GPT-4 to score the predicted answers from 0 to 10 and provide different scoring examples as a reference for GPT-4.

\subsubsection{Implementation details}
We set the number of analogical examples \textit{k} as 3. The size of the reasoning library $\mathcal{N}_{RL}$ is 100. In the fine-tuning stage, ChatGLM2-6B-32K and Baichuan2-13B-Chat are tuned with LoRA \cite{hu2021lora} on 1 and 4 A100 (80G). We set the optimizer, maximum context size, batch size, and learning rate to Adam, 4096, 8, and 5e-5 respectively. For testing, a single 80G A100 is used. We use the code framework from LLaMA-Factory\footnote{\url{https://github.com/hiyouga/LLaMA-Factory}} \cite{llama-factory}.
\begin{table*}[t]
    \centering
    \caption{NL2SELL results on GPT-3.5. The best results are in bold, while the second are underlined.}
    \begin{tabular}{l|>{\columncolor{light-gray}}c|c c c | c c c c |c }
    \toprule
     & {Overall Acc.}
     & \multicolumn{3}{c|}{Structure Acc.} 
     & \multicolumn{4}{c|}{Human Eval.}
     & {GPT4 Eval.}
     \\
     \cline{2-10} 
     & S-BLEU & L & R/O & Mean & Key & Value & Logic & Overall & GPT-S\\
     \hline 
      Zero-shot &  {37.0} &  {0.225} &  {0.738} &  {0.482} &  {0.549} &  {0.638} &  {0.597} &  {0.628} &  {5.87} \\
      Fixed few-shot &  {42.0} &  {0.368} &  {0.719} &  {0.543} &  {0.613} &  {0.692} &  {0.637} &  {0.735} &  {6.54}\\
      Fixed few-shot + RS  &  {52.8} &  {0.657} &  {0.771} &  {0.714} &  \underline{0.635} &  {0.701} &  {0.654} &  {0.734}  & {6.65} \\
      Random few-shot + RS & \underline{56.5} &  \underline{0.727} &  \underline{0.791} &  \underline{0.759} &  {0.626} &  \textbf{0.731} &  \underline{0.670} &  \underline{0.772} & \underline{6.74} \\
      \hline
      ARAP & \textbf{58.5} & \textbf{0.779} & \textbf{0.809} & \textbf{0.794} & \textbf{0.653} & \underline{0.715} & \textbf{0.678} & \textbf{0.774} & \textbf{6.84} \\
    \bottomrule
    \end{tabular}
    \label{tab:chatgpt_results}
\end{table*}
\subsection{Analogical Prompting result on GPT-3.5}
\subsubsection{Main Results}
Table \ref{tab:chatgpt_results} displays the results of different prompting methods on GPT-3.5, where ARAP represents our Analogical Reasoning Augmented Prompting. The experimental results demonstrate that: \textbf{(1)} ARAP significantly outperforms other prompting methods in various evaluation metrics, especially in terms of structural accuracy which improves by over 7\%, proving that analogical examples provide a good logical structure for reasoning. \textbf{(2)} Even randomly selected few-shot reasoning examples are better than fixed ones. This indicates that fixed reasoning examples for all inputs are often suboptimal, demonstrating the necessity of establishing a reasoning library to provide diverse reasoning samples. \textbf{(3)} Reasoning steps are essential in prompting. When comparing results between ``Fixed few-shot'' and ``Fixed few-shot + RS'', we found that reasoning steps have a significant active impact on reasoning accuracy, which is similar to the trend in other previous research works \cite{wei2022chain}. 
\begin{figure}[]
    \centering
    \includegraphics[width=0.5\textwidth]{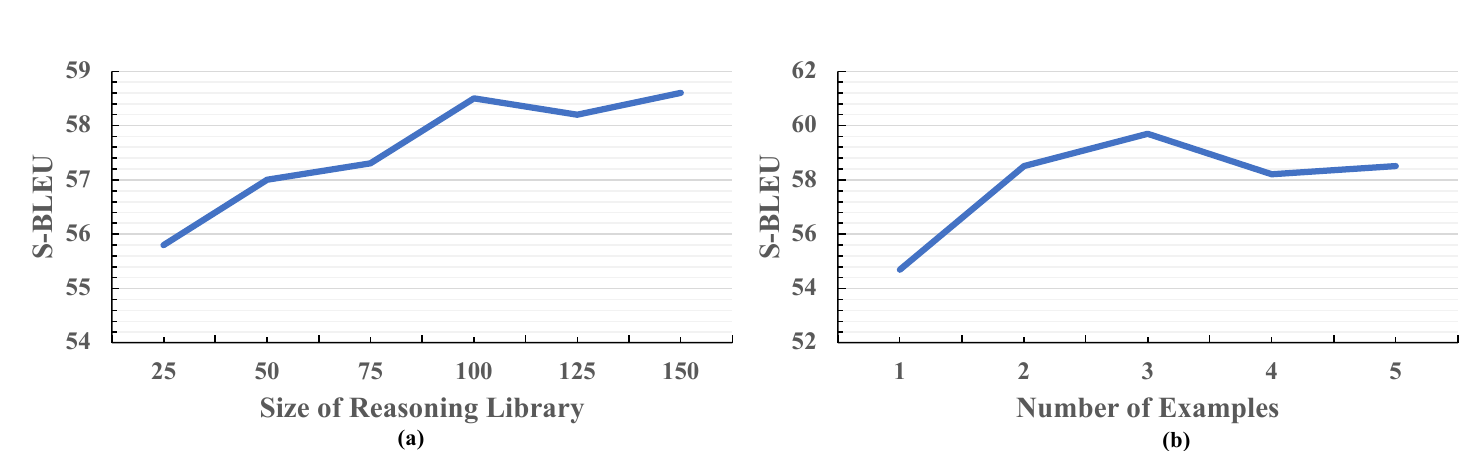}
    \caption{The impact of the size of parameter settings.}
    \label{analysis}
    \vspace{-1mm}
\end{figure}

Further, we explore the impact of parameter settings, such as the size of reasoning library $\mathcal{N}_{RL}$ and the number of analogical examples $k$. As shown in Figure \ref{analysis} (a), as the size of the reasoning library increases, the overall accuracy gradually increases, but the rate slows down, indicating that when the inference library reaches a certain scale, it can already provide references for most marketer's demands. In Figure \ref{analysis} (b), it can be observed that the overall accuracy first increases and then decreases when the number of examples in prompts increases. This reflects that when engaging in context-based learning, providing more examples to LLMs is not always better. When an excessive number of examples are input, irrelevant examples may introduce noise, and an overly lengthy context can also impose a greater burden on the model's understanding.

\subsubsection{Case Study}
\begin{table}[t]
    \centering
    \caption{Case study on GPT-3.5 with different versions of prompt. Bad responses are marked with a wavy underline.}
    \small
    \begin{tabularx}{0.45\textwidth}{l|X}
    
    \toprule
     \rowcolor{light-gray}\multicolumn{2}{c}{Demand:\textit{Company white-collar workers who enjoy drinking Starbucks}} 
     \\
      \hline
      Zero-shot &  {(Preference\#Belongs To\#Starbucks)}\\
      \hline
      Fixed few-shot  &  {(Preference\#Belongs To\#Starbucks) AND (Resident City\#Belongs To\#\uwave{Company white-collar)}}\\
      \hline
      Fixed few-shot + RS  &  {(Preference\#Belongs To\#\uwave{Enjoy} drinking Starbucks) AND (Resident City\#Belongs To\#\uwave{Company white-collar})} \\
      \hline
      Random few-shot + RS &  {(Preference\#Belongs To\#Starbucks)}\\
      \hline
      ARAP & \textbf{(Preference\#Belongs To\#Starbucks) AND (Career\#Belongs To\#White-collar)} \\
    \bottomrule
    \end{tabularx}
    \label{tab:case_study_1}
\end{table}

Table \ref{tab:case_study_1} is a case we obtained from GPT-3.5 with different versions of prompts. Although we provide the same tag list for a specific demand across different prompt versions, there are significant variations in the selection of key-value pairs. For the first demand ``Company white-collar workers who enjoy drinking Starbucks'', all baseline versions ignore the tag of ``Career'', indicating that the model cannot smoothly map the description ``white-collar workers'' to ``Career'', although this is common sense. Meanwhile, in both fixed few-shot versions, the model treats ``company white-collar workers'' as the value of the ``Resident City'', which is due to the answers of fixed examples in prompts containing similar expressions such as (\textit{Resident city}\#\textit{Belongs To}\#\textit{Hangzhou}), resulting in the model being inappropriately imitated. While in ARAP, we retrieve similar demands like ``{Young people who enjoy swimming or female white-collar workers who enjoy reading in Shanghai}'' and its reasoning steps from the reasoning library, thus helping the LLM analogically transform the ``white-collar workers'' into conditional expression ``(\textit{Career}\#\textit{Belongs To}\#\textit{White-collar})'' as they share the similar description ``white-collar workers''.
\subsection{Knowledge distillation results on smaller LLMs}
\begin{table*}[t]
    \centering
    \caption{NL2SELL results on distilled model Baichuan2-13B-Chat and ChatGLM-6B-32K. The best tuning results of each fine-tuned LLMs are in bold.}
    \small
    \begin{tabular}{c|l|>{\columncolor{light-gray}}c|c c c |c c c c |c }
    \toprule
     \multicolumn{2}{c|}{}
     & {Overall Acc.}
     & \multicolumn{3}{c|}{Structure Acc.} 
     & \multicolumn{4}{c|}{Human Eval.}
     & {GPT4 Eval.}
     \\
     \cline{3-11} 
     \multicolumn{2}{c|}{} & S-BLEU & L & R/O & Mean & Key & Value & Logic & Overall & GPT-S\\
     \hline 
     \textit{GPT-3.5} & \textit{ARAP} & \textit{59.7} & \textit{0.779} & \textit{0.809} & \textit{0.794} & \textit{0.815} & \textit{0.715} & \textit{0.678} & \textit{0.774} & \textit{6.84}\\
     \hline
      \multirow{4}{*}{Baichuan2-13B-Chat}
      & ARAFT & \textbf{62.2} &  \textbf{0.822} & \textbf{0.806} & \textbf{0.814} & \textbf{0.779} & \textbf{0.737} & \textbf{0.813} & \textbf{0.735} & \textbf{7.10} \\
      & \quad -  MT &  {60.8} &  {0.809} &  {0.794} &  {0.802} &  {0.754} &  {0.731} &  {0.800} &  {0.718} &  6.98\\
      & \quad -  RS &  {50.1} &  {0.741} &  {0.762} &  {0.751} &  {0.713} &  {0.661} &  {0.754} &  {0.679}  & 6.63 \\
      & Normal  & {54.9} & {0.812} &  {0.786} &  {0.806} & {0.754} & {0.712} & {0.809} & {0.715} & 6.77 \\
     \hline 
      \multirow{4}{*}{ChatGLM2-6B-32K}
      & ARAFT & \textbf{56.0} &  \textbf{0.829} & \textbf{0.773} & \textbf{0.801} & \textbf{0.740} & \textbf{0.715} & \textbf{0.790} & \textbf{0.710} & \textbf{7.13} \\
      & \quad -  MT &  {55.3} &  {0.784} &  {0.757} &  {0.770} &  {0.726} &  {0.689} &  {0.768} &  {0.686} &  {6.65}\\
      & \quad -  RS &  55.0 & {0.775} &  {0.742} &  {0.758} &  0.688 &  0.669 & 0.741 &  0.668  & 6.33 \\
      & Normal & {42.0} & {0.368} &  {0.719} &  {0.543} & {0.641} &  {0.603} &  {0.698} &  {0.612} &  {5.72} \\
    \bottomrule
    \end{tabular}
    \label{tab:lora_finetue}
\end{table*}
\subsubsection{Main Results}
Table \ref{tab:lora_finetue} shows the testing results on fine-tuned LLMs with the knowledge distilled from GPT-3.5, where ARAFT represents fine-tuning (FT) in the reasoning-augmented multi-task paradigm mentioned in section \ref{fine-tune}. All testing prompts are the same as ARAP prompts in GPT-3.5. As is shown: \textbf{(1)} Both two models show comparable capability to GPT-3.5, especially on Baichuan2-13B-Chat, where the score of S-BLEU is improved by over 5\% compared to GPT-3.5 after fine-tuning. This demonstrates the superiority of our distillation and training methods. \textbf{(2)} Compared with the results of the single-task training strategy (-MT), the multi-task training strategy brings improvement to the reasoning performance of the LLMs. Although during testing, we limit the model to only output the final answers without reasoning steps, adding training tasks that predict reasoning steps can enhance the models' reasoning abilities and robustness. \textbf{(3)} Compared with the results without reasoning steps in the input corpus (-RS), we can conclude that providing explicit reasoning steps in training input is beneficial. If there are no reasoning steps in the input corpus but a multi-task training strategy is used, it will increase the difficulty of the training task and lead to poor performance, which is particularly evident in Baichuan2-13B-Chat. \textbf{(4)} The result of fine-tuning using only the normal demand and answer pairs is not satisfactory, which indicates that it is challenging for LLMs to learn the data patterns solely based on the Q\&A pairs in situations where tasks are difficult. Therefore, it is necessary to design appropriate demonstrations and training tasks to fine-tune the LLMs.
\begin{table*}[]
    \centering
    \caption{Ablation study of \textit{answer to demand} distillation approach on Baichuan2-13B-Chat.}
    \small
    \begin{tabular}{c|c|>{\columncolor{light-gray}}c| c c c | c c c c |c }
    \toprule
     & \multirow{2}{*}{Use \textit{\textbf{a2d}} ?}
     & {Overall Acc.}
     & \multicolumn{3}{c|}{Structure Acc.} 
     & \multicolumn{4}{c|}{Human Eval.}
     & {GPT4 Eval.}
     \\
     \cline{3-11} 
     &{}& S-BLEU & L & R/O & Mean & Key & Value & Logic & Overall & GPT-S\\
     \hline 
      \multirow{2}{*}{ARAFT} &\CheckmarkBold & \textbf{62.2} & \textbf{0.822} &  \textbf{0.806} &  \textbf{0.814} &  \textbf{0.779} & \textbf{0.737} & \textbf{0.813} & \textbf{0.735} & \textbf{7.10} \\
        & \XSolidBrush &  {58.9} & {0.811} &  {0.784} &  {0.798} &  {0.752} &  {0.713} &  {0.790} &  {0.712} &  {7.01}\\
      \hline
      \multirow{2}{*}{ARAFT - RS} & \CheckmarkBold & \textbf{50.1} & \textbf{0.741} & \textbf{0.762} &  \textbf{0.751} & \textbf{0.713} & \textbf{0.661} & \textbf{0.754} & \textbf{0.679} & \textbf{6.63} \\
        & \XSolidBrush & {46.1} & {0.667} &  {0.728} &  {0.697} & {0.674} & {0.625} &  {0.700} &  {0.637} &  {6.37}\\
    \bottomrule
    \end{tabular}
    \label{tab:a2q_ablation}
\end{table*}
\subsubsection{Ablation Study}
As mentioned in section \ref{distill}, there are two sources of distilled knowledge we use when fine-tuning, \ie knowledge distilled from \textit{demand to answer} and \textit{answer to demand} approach. The knowledge distilled from \textit{demand to answer} is the most commonly used knowledge distillation approach, and what we mainly want to explore is the effectiveness of the knowledge distilled from \textit{answer to demand}. To verify this, we conduct ablation experiments on the best-performing models ``Baichuan2-13B-Chat'' with two training settings (\ie ARAFT and ARAFT - RS). Table \ref{tab:a2q_ablation} shows the effect of the knowledge distilled from \textit{answer to demand} (\textit{a2d}), it can be found that when removing this part of data during the fine-tuning, the performance on both ARAFT and ARAFT - RS significantly decreases, although this part of data only accounts for about one-tenth of the training data. This indicates that adding samples with a more uniform distribution of logic and tags to the training corpus can improve the robustness of the LLMs, and our distillation method from \textit{answers to demand} achieves this.
\section{Application}
\textbf{Model Deployment.} 
The LLM tuned in section 3.3 has been deployed online for application using an A10 with a memory capacity of 24G. To achieve optimal online inference performance, the deployment involves a total of 8 cards, including both the pre-production and online environments. Meanwhile, the retriever service based on BGE has also been deployed on an A10 GPU.

\begin{figure}[t]
    \centering
    \includegraphics[width=0.5\textwidth]{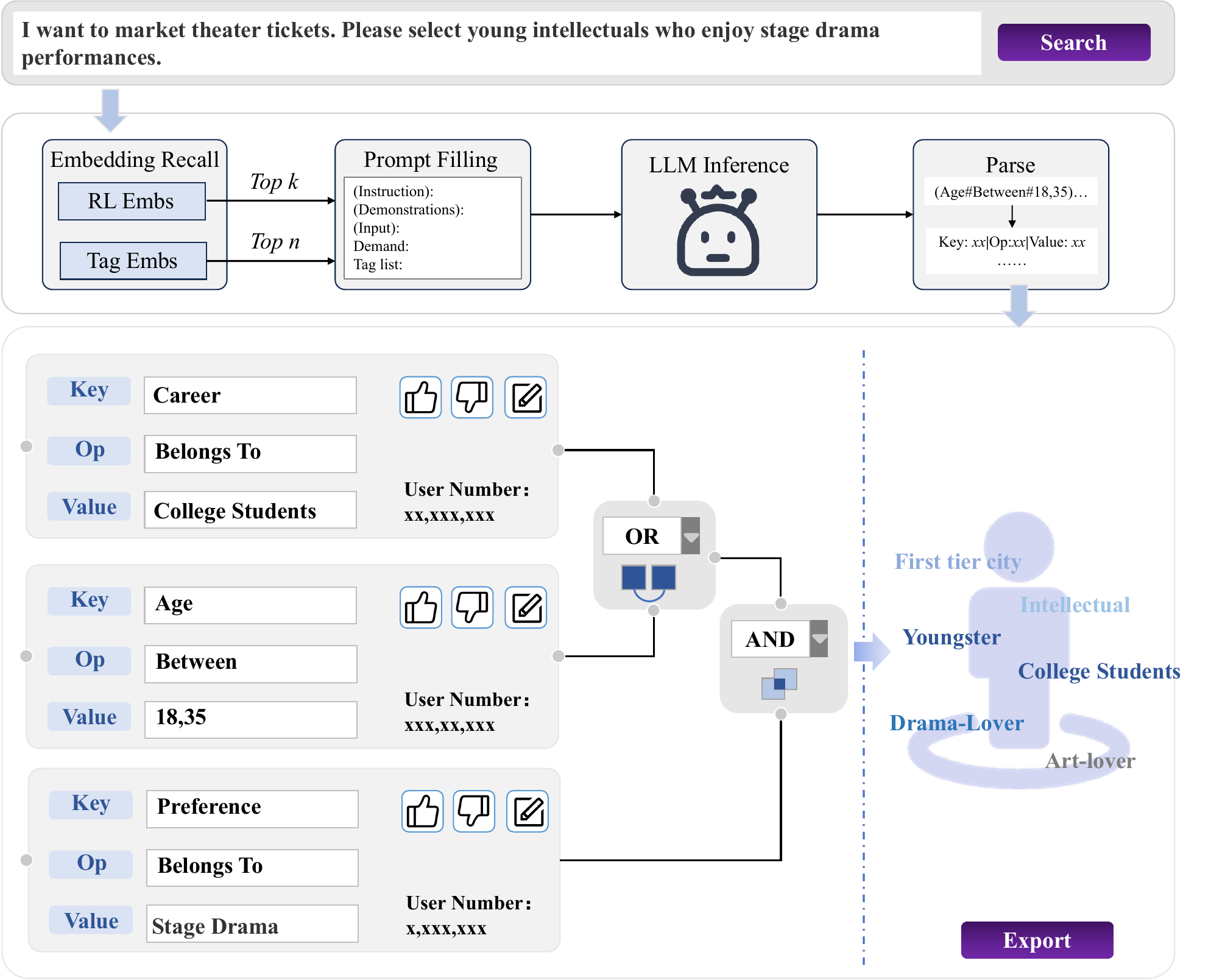}
    \caption{The application of ARALLM in User Targeting.}
    \label{application}
\end{figure}

\textbf{Application Case.} 
Figure \ref{application} shows an application case of ARALLM in online user targeting. A marketer wants to do marketing for a stage drama performance, so he can input a raw demand and click the ``Search'' button. Our system will invoke the retriever service to retrieve similar top-k demonstrations and top-n tags from the reasoning library embeddings (RL Embs) and tag embeddings (Tag Embs), respectively. After filling the prompt based on retrieved demonstrations and tags, our system requests the deployed ARAFT model to generate the answer expressed in SELL. The SELL expression will be parsed by our parser and results will be visualized to the marketers on the panel as shown in Figure \ref{application}. The marketers can verify and edit the card on the panel after clicking the edit button. Finally, they can click the ``Export'' button to get the target users. 

\textbf{Practical Effects.} Our system has been already running for months. We have set two metrics to evaluate the new approach online, including \textit{operation time}, and \textit{number of likes} from marketers. By collecting the operation log, the entire system takes no more than 10 seconds to complete a request and the average marketers' operation time of the new system is about 3 minutes, which is 4-10 times shorter than the former on average. Beyond that, we receive likes 1.3 times larger than the former, revealing the operation friendliness of this new way of user targeting.
\begin{table}[t]
  \caption{Online A/B testing results.}
  \label{ctr}
  \begin{tabular}{l|c|c}
    \toprule
     Campaign & CTR & Exposure\\
    \midrule
    Digital car & +55.21\% & +0.33\% \\
    Housing provident fund & +30.56\% & -0.21\% \\
    Ophthalmic health & +69.17\% & +0.01\% \\
    Traditional Chinese Medicine & +112.7\% & +0.67\% \\
    Recharge & +73.89\% & +0.20\% \\
  \bottomrule
  \end{tabular}
\end{table}

\textbf{Online A/B testing results.} We conduct online A/B experiments to compare our methods with the traditional rule-based methods and use the direct metrics CTR (Click-Through-Rate) and Exposure (the number of users who have been exposed by the campaign) to evaluate them. Results are shown in Table \ref{ctr}. From an end-to-end perspective, our solution has achieved better marketing results than rule-based methods in multiple marketing campaigns.

\section{Related Work}
\subsection{User targeting}
Methods of user targeting mainly can be categorized into two lines: rule-based and model-based methods. The rule-based methods \cite{mangalampalli2011feature,shen2015effective} match potential users with specific demographic tags (age, gender, geography) or interests that are targeted by marketers which need marketers to do user profile association or mining. The model-based methods expand a given seed set by calculating the similarity of all pairs between seed users and candidate users \cite{ma2016score,liu2016audience} or training customized prediction models for each service or campaign \cite{zhuang2020hubble,zhu2021learning}. These methods show good performance in user targeting but need seed sets. 
Both methods have primarily focused on intricate architectures for effectively and efficiently targeting users, while commonly disregarding the natural and significant gap between marketers' demands and the capabilities of current models, which can be time-consuming and unfriendly.

\subsection{NL2SQL}
Our NL2SELL task is somewhat similar to the task of translating natural language into SQL (NL2SQL).

NL2SQL is an important and challenging task in helping non-expert users to better manipulate databases. Graph neural networks (GNNs) are popular in NL2SQL research \cite{Wang_Shin_Liu_Polozov_Richardson_2020, Cao_Chen_Chen_Zhao_Zhu_Yu_2021, Hui_Geng_Ren_Li_Li_Sun_Huang_Si_Zhu_Zhu_2021}, which focus on modeling the relationship between the question and the schema in database system. Recently, large pre-trained language models (PLMs), such as T5 \cite{T5} and GPT-3 \cite{GPT3}, have shown strong translating ability in NL2SQL tasks. Due to the extensive knowledge injection during the pretrain phase, PLM-based methods achieve better results compared to GNN-based methods by fine-tuning them with a small amount of data. Some works also explore solving NL2SQL tasks with in-context learning \cite{Liu_Hu_Wen_Yu_2023,gu2023few}, it has been pointed out that LLMs show strong few-shot or zero-shot abilities even without any training data, which start a new direction for future research on NL2SQL field. The LLM-based methods in NL2SQL bring inspiration to our works.

\section{Conclusion and future work}
In this paper, we provide a novel user targeting approach by leveraging LLMs to gain a structured understanding of marketers' demands. We first define a new language SELL to express the needs of marketers more clearly. Subsequently, we propose an analogical reasoning augmented framework, ARALLM, which consists of analogical reasoning based prompting and reasoning-augmented multi-task model distillation. The experimental results on GPT-3.5 show that our analogical reasoning based prompting significantly outperforms other baseline prompting methods in the NL2SELL task. In addition, we distill a large-scale dataset using GPT-3.5 and train the student LLMs using a multi-task training approach, which is successfully used for online deployment. 

Since we primarily focus on the NL2SELL task in practical application scenarios, we have not yet applied the ARALLM framework to other reasoning tasks, such as coding or other structured language translation(\eg HTML, JSON), which will be explored as future work. At the same time, there are still labor costs in the construction of reasoning libraries and SELL datasets, we will explore more automated construction methods in the future. 

\begin{acks}
This work is funded by NSFC62306276, NSFCU23B2055 and NSFCU19B2027. This work is supported by the Fundamental Research Funds for the Central Universities (226-2023-00138). This work was supported by Ant Group. 
\end{acks}
 

\bibliographystyle{ACM-Reference-Format}
\bibliography{reference}

\appendix

\section{Design of SELL}
We provide a detailed explanation of the design and the usage of the {SELL}. The SELL is mainly composed of four elements:

\textbf{Keys} are a series of tags that describe the features of the user, such as \textit{Gender}, \textit{Marital Status}, \textit{Resident City}, and so on. In practical marketing scenarios, the database contains tens of thousands of tags that describe user profiles. Selecting the appropriate keys from this vast array of tags is a challenging task.

\textbf{Values} are the fillings of the corresponding keys. They can be generally divided into three types: numerical type, string type, and boolean type. For instance, the value corresponding to the key \textit{Monthly Income} is of numerical type, because they are continuous and non-enumerable. The value of \textit{Gender} is of string type, with limited options: \textit{Male} or \textit{Female}. Boolean type values have only two states, \ie \textit{True} or \textit{False}, such as \textit{Marital Status}. 

\textbf{Operators} represent the relationship between keys and values. For numerical values, seven types of operators can be used to connect them with corresponding keys: \textit{Equal To, Greater Than, Less Than, Not Equal To, Not Greater Than, Not Less Than,} and \textit{Between}. For values of string type and boolean type, there are two types of operators: \textit{Belongs To} and \textit{Not Belongs To}. Keys, operators, and values can form a basic \textbf{conditional expression} in {SELL}, formatted as ``(\textit{key\#operator\#value})'', such as ``(\textit{Marital Status\#Belongs To\#True})''. A conditional expression represents a cluster of target users, for example, in the above example, the target users are married. 

\textbf{Intersection and Union symbols} are used to take the intersection or union of multiple conditional expressions. In practical marketing scenarios, the features of the target users are often complex and difficult to describe using a single conditional expression. Therefore, it is necessary to combine the target users through intersection or union operations. In {SELL}, we use the symbols \textit{AND} and \textit{OR} to denote the intersection and union operations, respectively. For example, ``((\textit{Resident City}\#\textit{Belongs To}\#\textit{City A}) \textit{OR} (\textit{Resident City\#Belongs To\#City B})) \textit{AND} (\textit{Pet Owning\#Belongs To\#True})'' describes the pet owners who live in City A or City B.

Some demands and their corresponding SELL expressions are shown in Table \ref{tab:sell_ex}. More data can be found at \url{https://github.com/alipay/Analogic-Reasoning-Augmented-Large-Language-Model}.

\section{Human-Written Reasoning Steps}
As mentioned in the main paper, we write reasoning steps for 10\% of data in the reasoning library. Table \ref{tab:reasoning_ex} shows an example.

\section{Reasoning Steps Generation Prompt}
As mentioned in the main paper, we use GPT-3.5 to help us complete the reasoning steps of remained data. Figure \ref{fig:prompt0} shows the prompt we use in this step.
\begin{figure}
    \centering
    \includegraphics[width=0.4\textwidth]{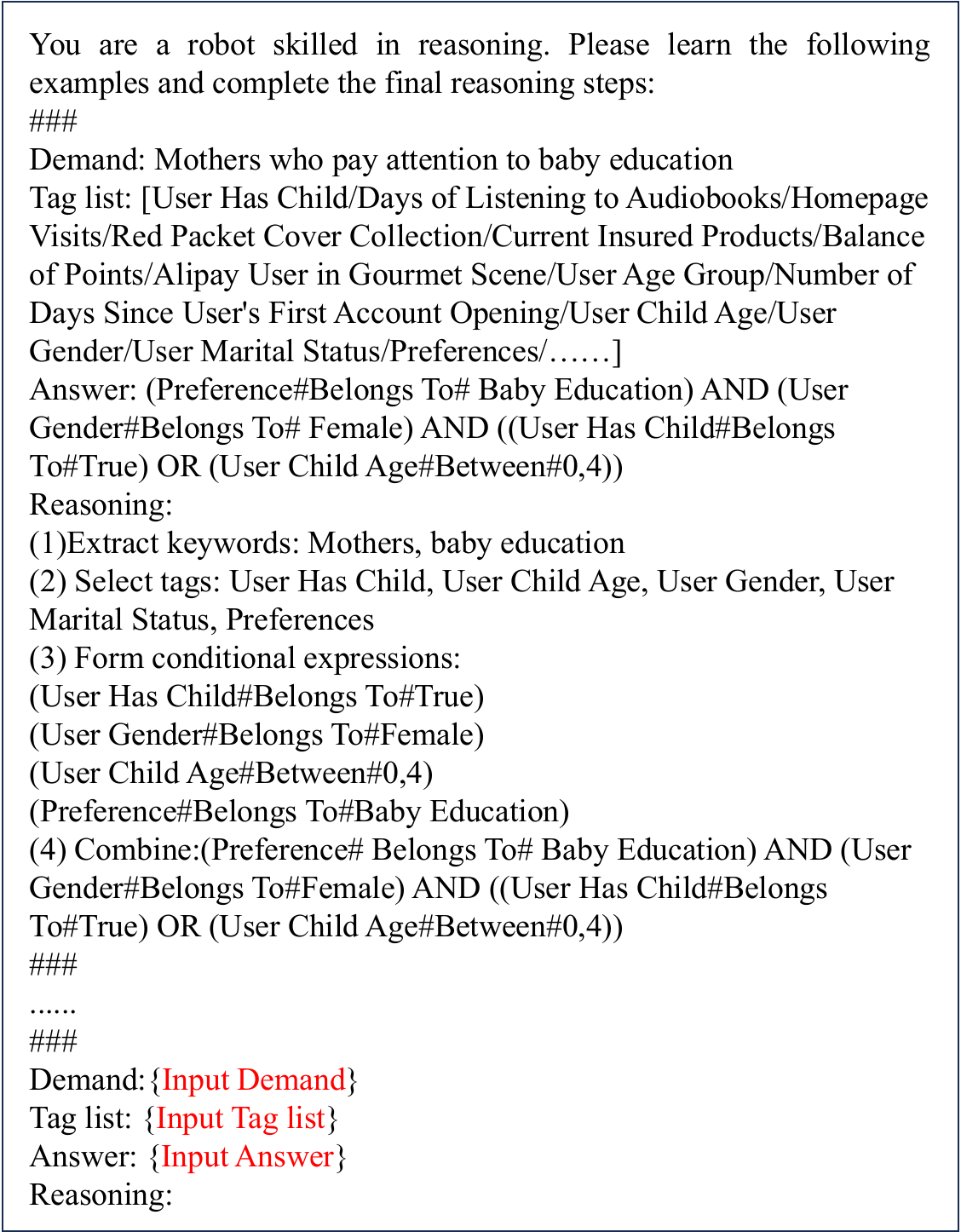}
    \vspace{-3mm}
    \caption{Reasoning steps generation prompt.}
    \label{fig:prompt0}
\end{figure}

\begin{figure}
    \centering
    \includegraphics[width=0.4\textwidth]{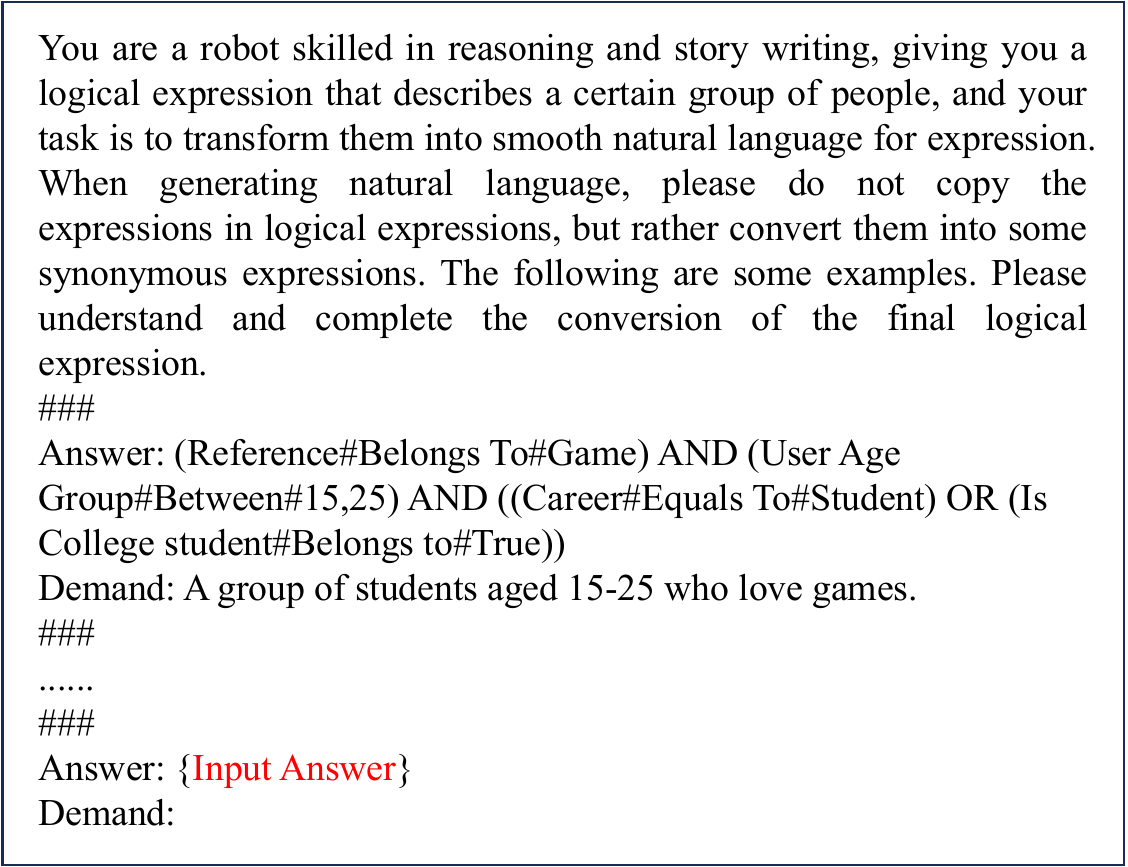}
    \vspace{-3mm}
    \caption{Demand generation prompt.}
    \label{fig:prompt1}
\end{figure}

\begin{table*}[t]
    \centering
    \caption{Some examples of marketers' demands and corresponding SELL.}
    \small
    \begin{tabularx}{\textwidth}{X|X}
    
      \toprule
      Middle-aged women &  {(User Age Group\#Between\#35,55) AND (Gender\#Belongs To\#Female)}\\
      \hline
      Young people in City A who enjoy swimming or female white-collar workers in City B who enjoy reading  &  ((Resident City\#Belongs To\#City A) AND (Preference\#Belongs To\#Swimming) AND (User Age Group\#Between\#18,35)) OR ((Resident City\#Belongs To\#City B) AND (Gender\#Belong To\#Female) AND (Career\#Belongs To\#White-collar) AND ((Preference\#Belongs To\#Reading) OR (Days of Listening To Audiobooks\#Greater Than\#1)))\\
      \hline
      Users who have a certain understanding or interest in financial products, insurance, etc.  &  (Preference\#Belongs To\#Wealth Management Products) OR (Preference\#Belongs To\#Insurance) OR (Eligible group for Wealth Infinity Card\#Belongs To\#True) OR (Has actively invested in major financial products\#Belongs To\#True) \\
     \bottomrule
    \end{tabularx}

    \label{tab:sell_ex}
\end{table*}
\begin{table*}[t]
    \centering
    \caption{Examples of human-written reasoning steps.}
    \small
    \begin{tabularx}{\textwidth}{l|X}
    
      \toprule
     {Demand} & {Mothers who pay attention to baby education.}\\
      \hline
      Tag List &  User Has Child/Days of Listening to Audiobooks/Homepage Visits/Red Packet Cover Collection/Current Insured Products/Balance of Points/Alipay User in Gourmet Scene/User Age Group/Number of Days Since User's First Account Opening/User Child Age/User Gender/User Marital Status/Preferences/......\\
      \hline
      Reasoning Steps &  \makecell[l]{\textbf{(1)Extract keywords:} Mothers, baby education \\ \textbf{(2) Select tags:} User Has Child, User Child Age, User Gender, User Marital Status, Preferences \\ \textbf{(3) Form conditional expressions:}\\(User Has Child\#Belongs To\#True)\\ (User Gender\#Belongs To\#Female)\\(User Child Age\#Between\#0,4)\\(Preference\#Belongs To\#Baby Education)\\ \textbf{(4) Combine:} \\(Preference\# Belongs To\# Baby Education) AND (User Gender\#Belongs To\# Female) AND ((User Has Child\#Belongs \\To\#True) OR (User Child Age\#Between\#0,4))} 
      \\
      \hline
      Answers &  (Preference\#Belongs To\#Baby Education) AND (User Gender\#Belongs to\#Female) AND ((User Has Child\#Belongs To\#True) OR (User Child Age\#Between\#0,4)) \\
     \bottomrule
    \end{tabularx}

    \label{tab:reasoning_ex}
\end{table*}

\begin{figure}
    \centering
    \includegraphics[width=0.4\textwidth]{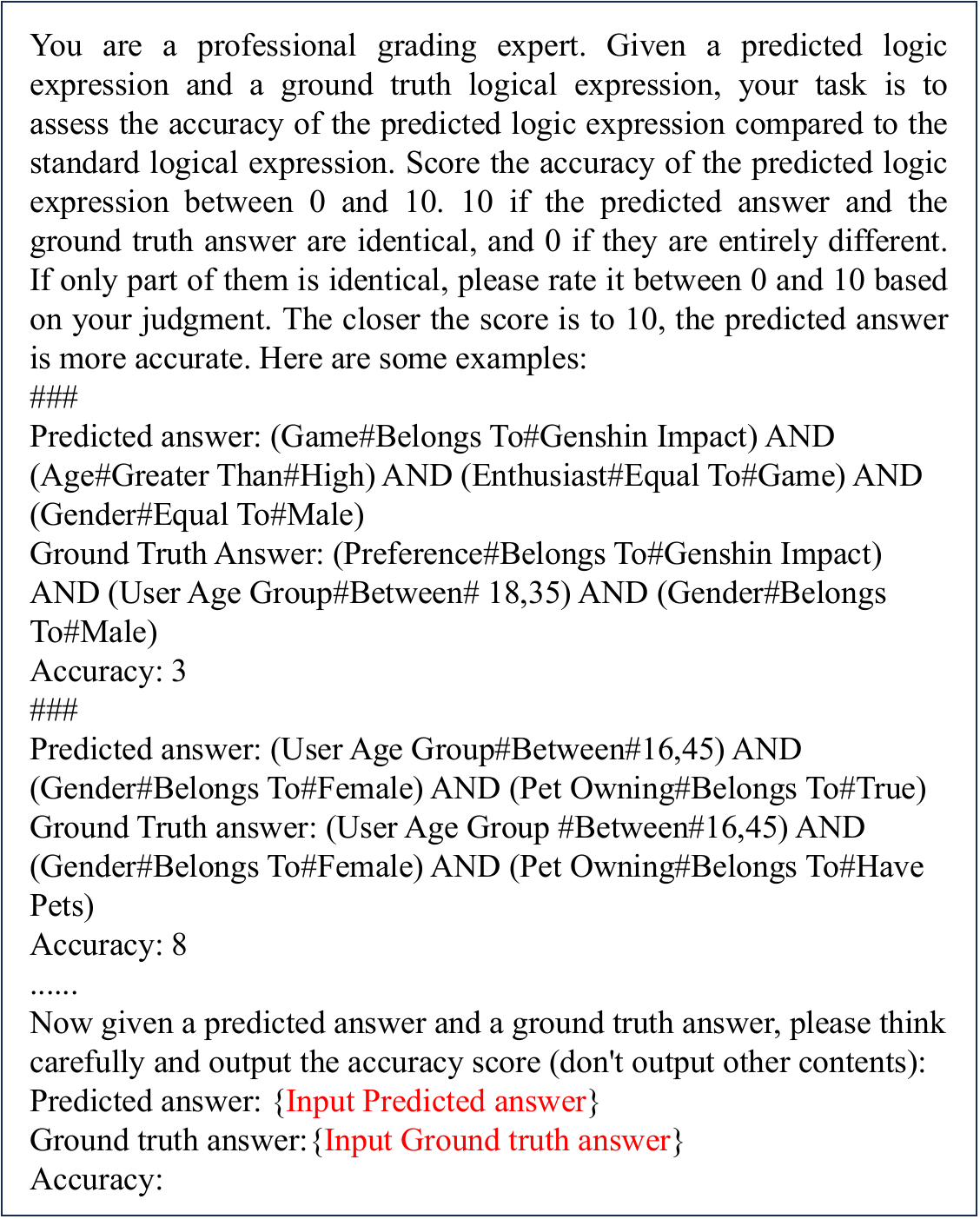}
    \vspace{-3mm}
    \caption{GPT4 Evaluation prompt.}
    \label{fig:prompt2}
\end{figure}

\begin{figure}
    \centering
    \includegraphics[width=0.4\textwidth]{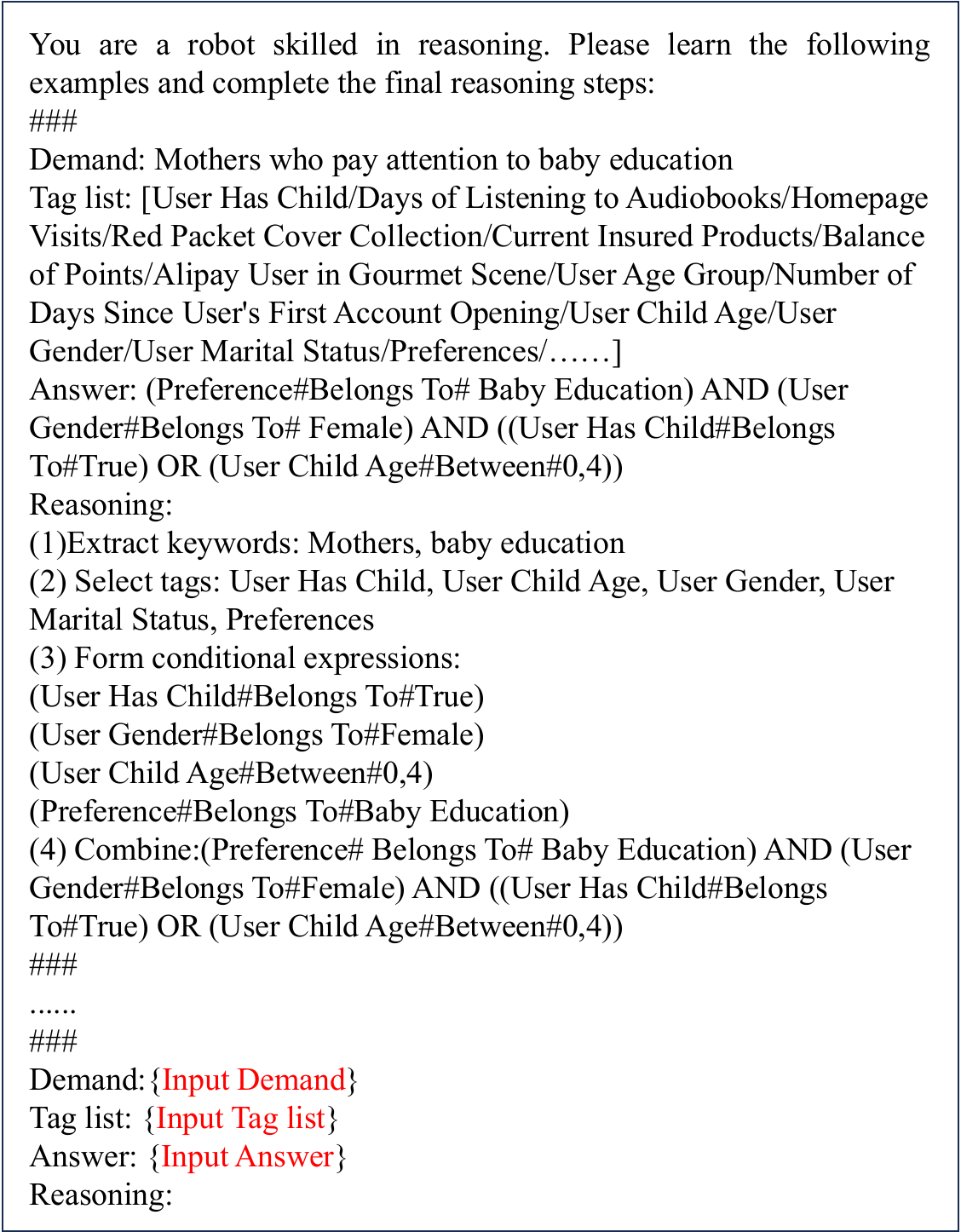}
    \vspace{-3mm}
    \caption{Reasoning steps generation prompt.}
    \label{fig:prompt0}
\end{figure}

\section{Demand generation prompt}\label{demand gen}
The prompt framework for demand generation given SELL expressions is shown in Figure \ref{fig:prompt1}.

\section{Logic Template}
As mentioned in the main paper, we first construct a series of logical templates of SELL to artificially synthesize answers. Table \ref{tab:template} shows all the logical templates we used. The \textit{c} in the template represents the basic conditional expression in SELL. To ensure that the synthesized answers are logically reasonable, we limit the number of conditional expressions in the logical template, and abandon overly complex intersection and union operations.
\begin{table}[b]
  \caption{Logic template for answer generation.}
  \small
  \begin{tabular}{c}
    \toprule
    \textbf{Template}\\
    \midrule
    \textit{c} \\
    \hline
    \textit{c} AND \textit{c}\\
    \hline
    \textit{c} AND \textit{c} AND \textit{c}\\
    \hline
    \textit{c} AND \textit{c} AND \textit{c} AND \textit{c}\\
    \hline
    \textit{c} OR \textit{c}\\
    \hline
    \textit{c} OR \textit{c} OR \textit{c}\\
    \hline
    \textit{c} OR \textit{c} OR \textit{c} OR \textit{c}\\
    \hline
    (\textit{c} AND \textit{c}) OR \textit{c}\\
    \hline
    (\textit{c} OR \textit{c}) AND \textit{c} \\
    \hline
    (\textit{c} AND \textit{c}) OR (\textit{c} AND \textit{c}) \\
    \hline
    (\textit{c} OR \textit{c}) AND (\textit{c} OR \textit{c})\\
    \hline
    \textit{c} AND (\textit{c} OR \textit{c} OR \textit{c})\\
    \hline
    \textit{c} OR (\textit{c} AND \textit{c} AND \textit{c})\\
    \hline
    (\textit{c} AND \textit{c}) OR (\textit{c} OR \textit{c})\\
    \hline
    (\textit{c} AND \textit{c}) AND (\textit{c} OR \textit{c})\\
    \hline
    \textit{c} AND ((\textit{c} AND \textit{c}) OR \textit{c})\\
    \hline
    \textit{c} AND ((\textit{c} OR \textit{c}) AND \textit{c})\\
    \hline
    \textit{c} OR ((\textit{c} OR \textit{c}) AND \textit{c})\\
    \hline
    \textit{c} OR ((\textit{c} AND \textit{c}) OR \textit{c})\\
    \bottomrule
  \end{tabular}
  \label{tab:template}
\end{table}
\section{GPT4 Evaluation Prompt}\label{gpt4 eval}
The prompt framework for guiding GPT4 to evaluate the quality of answer SELL is shown in Figure \ref{fig:prompt2}.
\end{document}